# Pedestrians play chicken with an autonomous vehicle

Rakshit Soni[1,2] and Charles Fox[1]

[1] School of Engineering and Physical Science, University of Lincoln, UK
chfox@lincoln.ac.uk
[2] CESI, CESI LINEACT, 92074 Paris La Défense, France
rsoni@cesi.fr

**Abstract.** Automated vehicles (AVs) are commonly programmed to yield unconditionally to pedestrians in the interest of safety. However, this design choice can give rise to the Freezing Robot Problem in which pedestrians learn to assert priority at every interaction, causing vehicles to stall and make no progress. The game theoretic Sequential Chicken model has shown that, like human drivers, AVs can resolve this problem by trading credible threats of very small risks of collision or larger risks of less severe invasion of personal space against the value of time due to yielding delays. This paper presents the first demonstration and evaluation of this approach using a real AV with human subjects and shows that pedestrian behavior under experimentally constrained safety conditions can be well fitted by Sequential Chicken, with a low time value of collision, suggestive of their planning to avoid proxemic personal space penalties as well as actual collisions.

## 1 Introduction

Interacting with human pedestrians in negotiations for shared space remains a challenge for autonomous vehicles [1]. The Freezing Robot Problem is the dilemma in which robots maneuvering in shared spaces may make little or no progress due to overly cautious safety algorithms [2]. If a robot is programmed to stop whenever an obstacle is in its path, human pedestrians quickly learn that they can push in front of it at any time [3]. The robot can then freeze as it continually yields to every pedestrian in a shared environment.

Sequential Chicken is a game-theoretic model proposed to solve the freezing robot problem [4]. It draws inspiration from the classic game of Chicken, in which two drivers approaching each other head-on must both decide whether to swerve or hold their course, risking a collision if neither chooses to yield[3]. Sequential Chicken treats the pedestrian and the AV as players in a sequence of discrete games, each making strategic decisions based on their own goals and beliefs about the other's behavior. Each player has defined utilities for the different outcomes: such as reaching their destination with different delays, or colliding. The model finds optimal strategies for both players, where each chooses the action either to yield or not, that maximizes their expected utility given the other's likely choices. The optimal strategies for both players are probabilistic, and gradually increase the probability of yielding as they get closer together, whilst always retaining a strictly non-zero, but sometimes very small, probability of non-yielding to create a credible threat. The model can be used to fit parameters to human behavioral data, and for prescriptive control of autonomous vehicles.

[3] Examples occur in movies *Rebel Without a Cause* and *Back to the Future Part II*

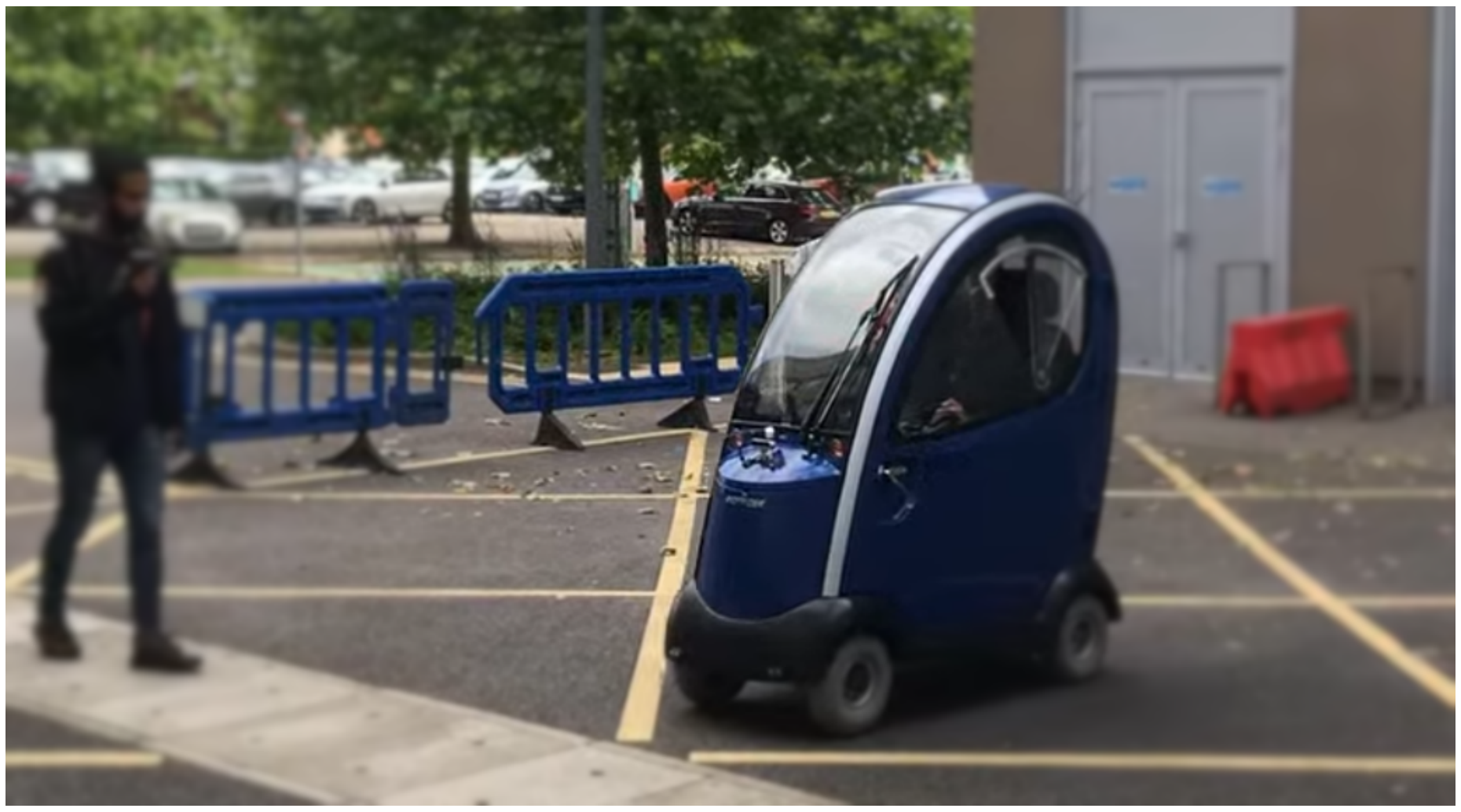

**Fig. 1.** Sequential Chicken experiment with OpenPodcar2

Empirical tests of Sequential Chicken between two human players were performed in lab-based discrete board games then physical versions of these board games [5]. Continuous physical setups were then performed [6]. Virtual Reality (VR) interactions between human subjects and large and small autonomous vehicles were further studied, which included the possibility of actual collisions [7] [8]. All of these studies could fit Sequential Chicken to the human behaviors, though the implied costs of collision has appeared consistently lower than would be expected in real life. This may be due to lack of realism of the artificial settings. Theoretical extensions suggest that actual collisions become unnecessary as credible threats if they can be replaced by much smaller but more frequent penalties such as invasion of personal space, which related models show can be generated and modeled quantitatively [9], [10], [11]. This provides a competing explanation for the observed human behaviors in lab and simulation studies: if vehicles are able to inflict immediate small psychological penalties onto pedes-trians simply by *almost* touching them, then we would expect to see a larger number of near-contacts but with a smaller implied negative utility than for actual collisions.

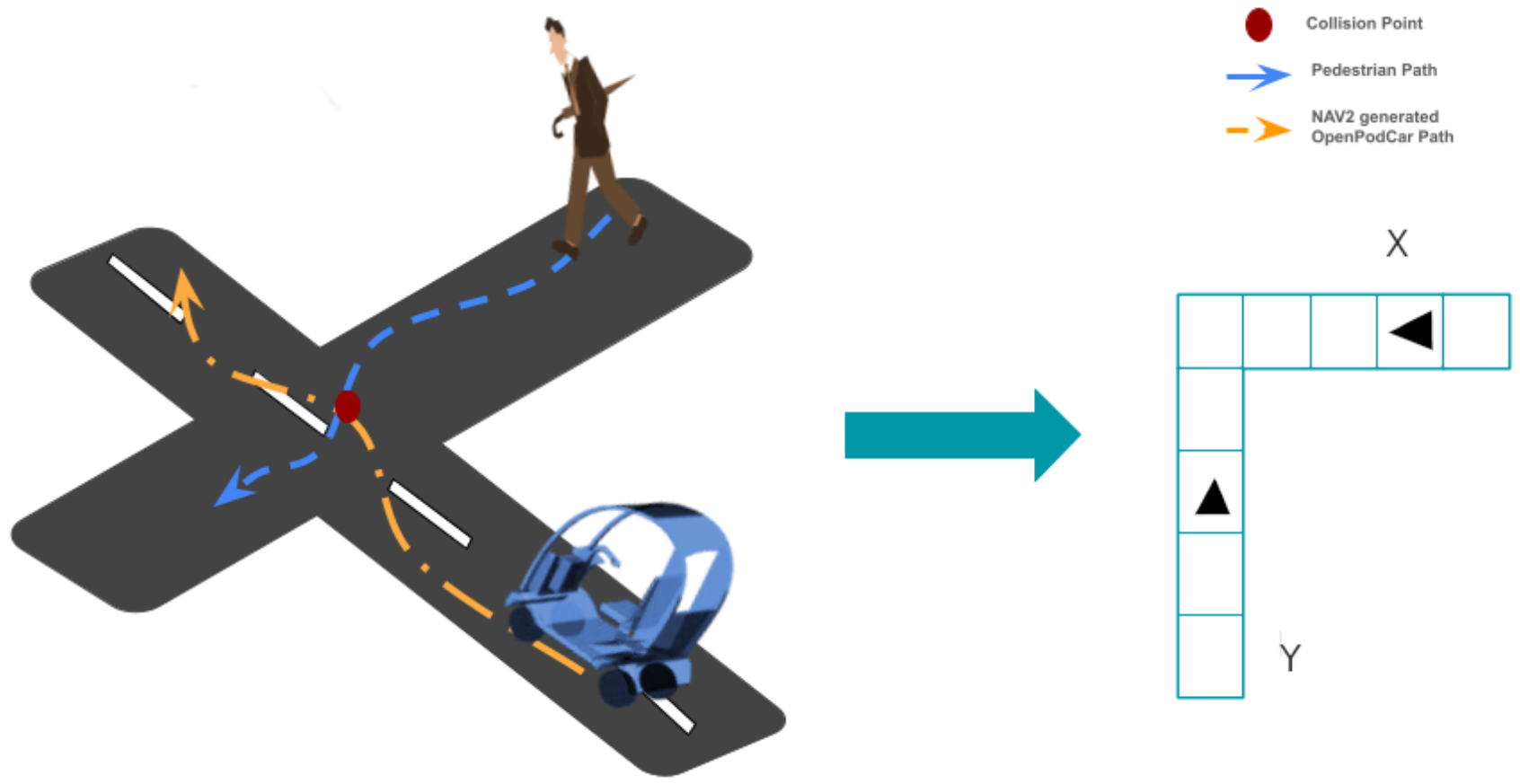

**Fig. 2.** Path quantization.

In contrast to prior studies, this work presents the first, proof-of-concept real-time fitting of Sequential Chicken model to human subjects interacting with a physical autonomous vehicle. Strict safety requirements constrain the experiment to a controlled environment, artificially reducing speeds of both agents; while modeling requirements restrict both to linear motion with no swerving. The results are used to assess whether the implied value of penalties in the best fit model align with those reported in previous studies.

### 1.1 Sequential Chicken model

Sequential Chicken [4] extends the classic Chicken model from game theory [12] to a series of games. Chicken is a single-turn game with a payoff matrix such as:

$$\begin{bmatrix} (0,0) & (1,-1) \\ (-1,1) & (-1000,-1000) \end{bmatrix} \tag{1}$$

where columns represent player X choosing to go slow or fast; and rows represent player Y choosing slow or fast; and matrix element pairs represent the utilities to (player X, player Y) of the result. This models the case of two agents choosing to yield to avoid collision where only a single immediate decision is available to determine the outcome.

In Sequential Chicken, a pedestrian $X$ and autonomous vehicle $Y$ approach each other, in space quantized into discrete boxes as in Fig. 2, and time quantized as discrete turns. On each turn, both agents simultaneously select from discrete speed actions: $SLOW = 1$ square per turn; or $FAST = 2$ squares per turn. The discrete positions of the players are represented by integers $(y, x)$ at discrete turns $t$, with their speed selection actions denoted $a_Y, a_X \in \{1, 2\}$. The new state at turn $t+1$ is thus $(y + a_Y, x + a_X)$. A key concept of the model is the use of time (seconds) as currency, to simplify modeling of value of travel time against risk of collision. The model asks how many seconds is a human willing to spend or save in exchange for avoiding another negative event; and thus, the time value of increasing and reducing the risk of that event by different amounts. This reduces the model's parameters, as it no longer needs to find utilities of both time and collisions in any other utility units.

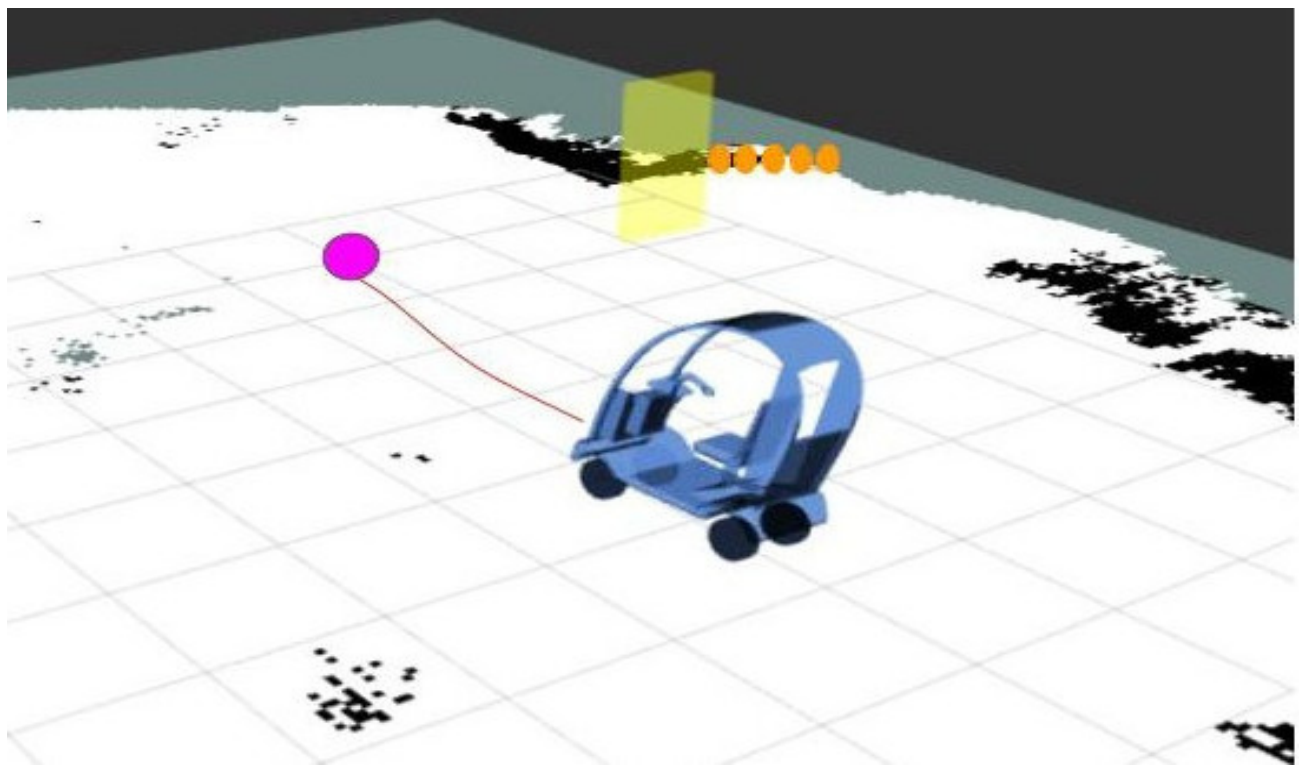

**Fig. 3.** Inferred world state in rviz: pedestrian (yellow) tracked path (orange circles); vehicle plan (red line); intersection point (magenta circle); and 2D SLAM map (black-/white ground).

Parameter $U_{crash}$ represents the (negative) utility of both agents occupying a box at or next to the collision point, $v(x \in \{0, 1\}) \wedge (y \in \{0, 1\}) = U_{crash}$. This may involve a physical contact collision or just an invasion of personal space within the box. Two boxes 0 and 1 are used in this definition so that the *FAST* action cannot jump over a single crash box – this would not be a good physical model of the agents passing through one another. Both agents aim to pass the collision point as quickly as possible to reach their destination and minimize the value of time spent on their journey. If an agent goes *SLOW* to yield and enable the other to pass first, this incurs additional cost of time. However, if neither yields, a collision occurs, resulting in the significantly larger negative utility, $U_{crash}$, for both.

The discretizations simplify the model, but by selecting sufficiently small squares and short time steps, the model can approximate the continuum of possible positions and speeds that lie between the quantized slow and fast speeds.

The model is intentionally designed to be simple and solvable, so excludes scenarios such as lateral steering around the other player or any form of communication other than the observable speed choices. It does not include actions for complete stops, only for *SLOW*, to ensure finite solutions. Assuming both players play optimally, the value $v_{y,x,t}$ of the subgame $(y, x, t)$ is defined as the pair of expected utilities for the two players, $(v^Y_{y,x,t}, v^X_{y,x,t})$. Following standard game theory principles, this can be written [4] as:

$$v_{y,x,t} = v\left(\begin{bmatrix} v_{y-1,x-1,t+1} & v_{y-1,x-2,t+1} \\ v_{y-2,x-1,t+1} & v_{y-2,x-2,t+1} \end{bmatrix}\right) \quad (2)$$

where the $2 \times 2$ matrix is the payoff matrix for the game, and the value of a payoff matrix is the expected utilities under its Nash equilibrium.[4] This recursive equation defines the value $v_{y,x,t}$ in terms of the values of the resulting next games' values, such as $v_{y-1,x-1,t+1}$. The recursion terminates as long as both agents actions are either *FAST* or *SLOW*, ensuring that they will eventually reach their destinations or collide, with either consequence immediately determining the value at that time in terms of numerical utility. Eqn. 2 is solved recursively [4].

As the players approach, in optimal solution the probability of each of them yielding over $t$ gradually increases towards 1/2, so the cumulative probability of having yielded by $t$ increases towards but never quite reaches 1. When one player yields without the other doing so on the same turn, the game symmetry is broken and the solution becomes deterministic with both players moving *FAST* for the rest of the interaction, and the other player passing first [4].

## 2 Methods

### 2.1 OpenPodcar2 vehicle

We implement Sequential Chicken control on OpenPodcar2 [13], an autonomous vehicle able to transport one human passenger, shown in Fig. 1. OpenPodcar2 is based on the open source hardware and software OpenPodcar [14], and modified using an OSMC motor driver [15] for speed control, and an R4 control board [16] providing a ROS2 [17] interface to control the steering and speed. Vehicle speed is limited in R4 firmware to 0.2m/s for safety. An RGBD depth-cam (Intel Realsense D435) is mounted on the front bonnet, and rotated 60 degrees to the right so its field of view covers pedestrian approaches from the right.

The vehicle outputs its location in ROS2 map frame meters – using RTAB-Map SLAM algorithm [18] from the depthcam – and tracked 3D pedestrian poses measured in map frame meters as BoundingBox3D messages defined in ROS2's standard vision_msgs library and computed using YOLOv8[5] with ByteTrack [19] from the same depth-cam images used for localization. On detection, pedestrian bounding boxes are masked from RGBD images to avoid them affecting localization and mapping.

The vehicle navigates autonomously using SMAC planner (a hybrid of global A* of and local Dynamic Window Band (DWB)) [20]. The passenger sets a goal location on a laptop, then holds down a firmware-level dead man's handle (DMH) during autonomous driving to this location. Releasing the DMH at any time immediately cuts power to the motors and stops the vehicle.

[4] The present study assumes symmetric utilities for the two players, which ensures [4] there is always exactly one mixed-strategy equilibrium.

[5] `github.com/Rak-r/yolov8_ros2_OpenPodCarV2`

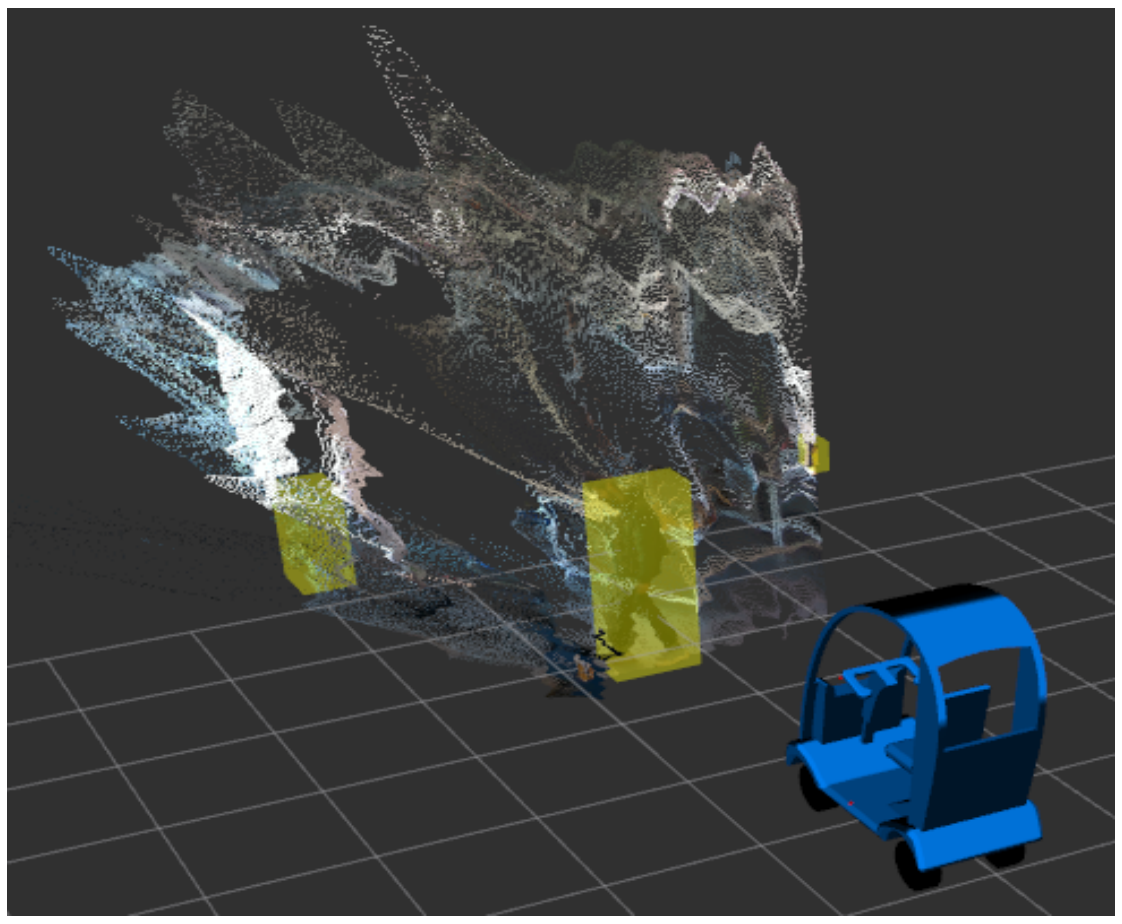

**Fig. 4.** Sensor inputs: 3D point cloud and detected pedestrian in front of OpenPodcar2, displayed in rviz.

### 2.2 Sequential chicken setup

Pedestrian boxes are chosen to be 0.2m, while car boxes are 0.08m, so that both agents cover about two boxes per second when *FAST*, assuming the safety speed-limited vehicle and reduced walking-speed subjects (adult human free walking speed is around 1.4 m/s [21] but subjects are slowed down by the experimental protocol in proportion to the car's safety speed limit, see Section 2.5).

### 2.3 Vehicle speed control

The vehicle is controlled to check for possible collision and to modulate its speed whenever a pedestrian is detected in front of the vehicle, as demonstrated in Fig. 4. The vehicle does not alter its planned path, only its trajectory.[6] The controller assumes there is at most one pedestrian. The pedestrian's trajectory is estimated, assuming they will move in a straight line (linear kinematic) in their current heading. The vehicle's own future path plan (which may be curved in general, but is linear in the present experimental settings) is available from nav2. Spatial intersection points of the two agent's paths are computed. The controller then finds the time it takes for each agent to arrive at the (first) intersection point, at their *FAST* speed. If these differ by at least the time taken for one to completely pass the other, then no game playing is needed. If they differ by less than this, the state will result is a collision at the current speeds and we define the state as *interesting*.

If the state is interesting, Sequential Chicken is played. The controller first quantize temporal progress along both paths into boxes. (This can include progress along spatially curved paths for the vehicle). These boxes are passed as input to the Sequential Chicken model which returns an immediate $FAST$ or $SLOW$ action for the vehicle, drawn from its optimal strategy. For $FAST$, the vehicle is set to travel at maximum speed. For $SLOW$ it is set to travel at half of this speed. The Sequential Chicken model and controller are Markovian; they do not make use of any history and each decision is make using only the current state. The pedestrian might change their heading so its useful to recalculate their straight line each time a new decision is to be made.

[6] A *path* is a sequence of poses, a *trajectory* adds pose timings to a path.

### 2.4 Software Implementation

We update our SMAC plan once per second. Pedestrians are masked from depth images on detection, so they do not appear in the map or costmaps. All plans thus pass through them, and re-planning does not move plans to drive around them. Plan updates are therefore minor, usually just in response to small controller errors or to SLAM updates. The controller server outputs cmd_vel messages.

Fig. 5 shows the new, open-source ChickenSpeedModulator ROS2 node created for the implementation, and which can now also be inserted into any nav2-compliant robot to add Sequential Chicken interaction.[7] ChickenSpeedModulator contains the Sequential Chicken game theory model implemented using the NashPy library [22], wrapped in the vehicle speed control logic implemented in Python3. On initialisation, it computes recursive solutions for the whole game, which takes several seconds. It then consults these solutions in real time for specific cases when needed online. $U_{crash} = 100s$ is used, based on results of previous studies [6], [7], [8].

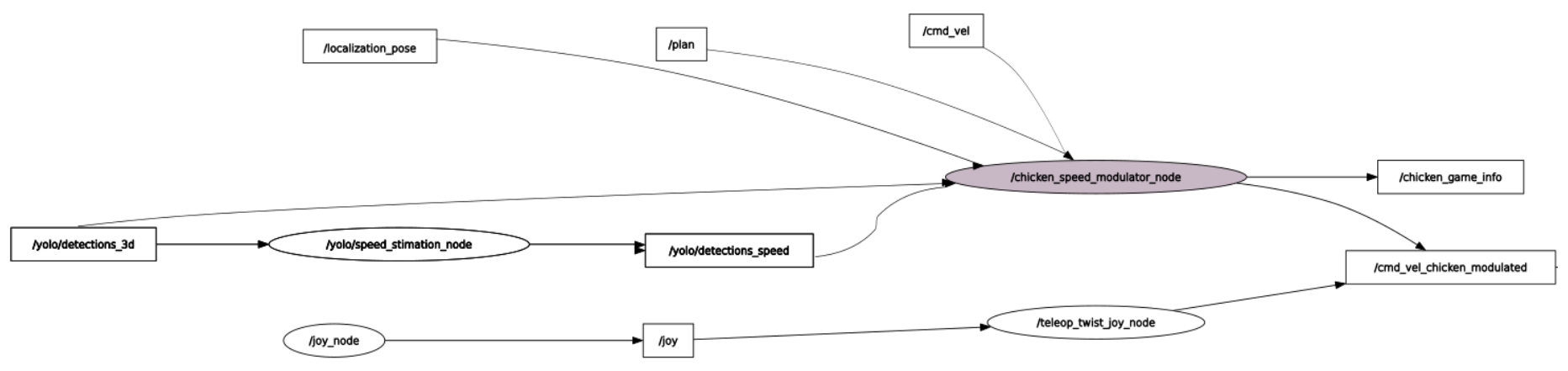

**Fig. 5.** ChickenSpeedModulator node, neighbours, and messages. It is open sourced so can be added to any nav2 setup.

All computations are in the map frame. The node subscribes to the robot pose ( localization_pose ) and pedestrian tracks (yolo/detections_speed; including pose and velocity), and to the planner-generated plan, in the map frame. The plan is a list of waypoint poses, to be followed at any speed.[8] The latest message on each topic is cached, and an update (step) function runs every 0.5 seconds. When a detection is present in an update, and the state is interesting, we play chicken. Each chicken speed decision is implemented for the next 0.5s as a modulation of the live cmd_vel messages, multiplying its original speed by 0.5 if a *SLOW* action is taken and leaving it unchanged for *FAST*. The game information is then published as custom ROS2 message over the topic chicken_game_info.

[7] `https://doi.org/10.5281/zenodo.19818131`

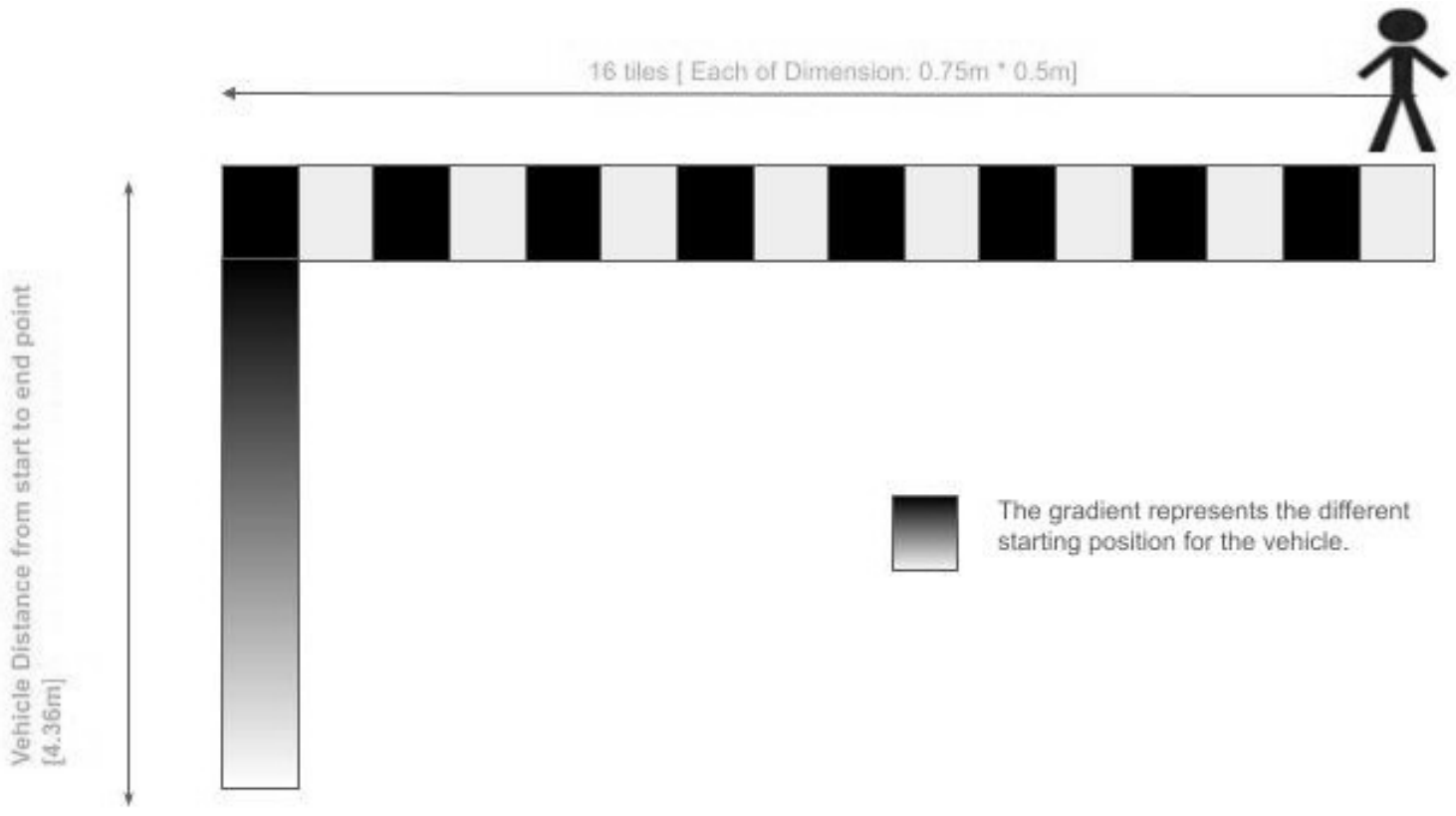

**Fig. 6.** Layout of the experiment

### 2.5 Human experimental protocol

Human experimental subjects were recruited as volunteers and signed consent under the University Research Ethics policy. Instructions were given to subjects as printed text, which they read at the start of the experiment:

> *Your task is to complete 20 road crossings from the start point to the end point, around a self-driving vehicle. You must walk using small steps so that every tile contains one foot. You must not run. Begin each crossing after the countdown. On completing a crossing, return to the start area for the next one. The vehicle uses AI to detect and interact with you. Like a human driver, it will change its speed to let you pass or to try to pass in front of you. You can watch the vehicle to see what it does and decide when you want to let it pass. A human safety passenger will sit in the vehicle. They will abort the experiment if your safety is at risk, then you will be considered to be 'dead' and your crossing failed. We will record your and the vehicle's motions. We will securely store your name, email, and consent, which will be used to send you a copy of the results but not linked to experimental data.*

[8] `docs.ros2.org/foxy/api/nav_msgs/msg/Path.html`

Fig. 1 shows the experiment. The pedestrian start area is 6-8m (12-15 tiles/-footsteps/boxes of 0.5m) away from the collision point. The initial vehicle start point is 4.3m away. The experimenter counts down to each crossing, and activates the vehicle when it begins by holding down the DMH. Position estimates of pedestrian and walking speed are recorded using the detection and tracking system on OpenPodcar2. Fig. 3 shows an approaching pedestrian with track history towards the end of the interaction with the vehicle while the geometric layout of the experiment has been demonstrated in fig 6.

Safety is critically important, as the vehicle must by law never actually hit pedestrians in experiments. Masking pedestrians (or any obstacles) from depth images presents a severe safety risk, as multiple layers of standard ROS2 safety systems all assume that obstacles can be seen in them and be used to prevent collisions. This severe risk is fully mitigated by the use of the firmware-level manual DMH cutoff in place of those systems. The instructions do not use deception. The game is cooperative so it is useful for pedestrians to have a true understanding of what the vehicle is doing. The requirement on subjects to walk with one foot placed in one tile is included to slow subjects down around 3.5 times from their usual walking speed, through a mixture of forcing smaller steps than normal, and forcing attention on the unusual walking. The vehicle speed is limited by our firmware to 0.2m/s for safety purposes, which is slower than most real world delivery vehicles, and this pedestrian requirement slows them down by a similar ratio to match.

The instructions say that the whole crossing will be aborted, rather than just the vehicle stopped, in the event of a risk. This is the closest outcome we can safely approximate the large negative utility from a real collision. The instructions are given as printed text in order to ensure that all subjects receive the exact same instructions. Verbal questions from subjects were referred to the text, with subjects allowed to interpret the text as they see fit. We assume that subjects value their time as in real life during the experiment. So requesting them to complete a fixed number of 20 crossings, rather than to collect data for a fixed time, motivates them to finish the experiment quickly and save their real-life time. The human experimenter (passenger) protocol is to adjust the starting location of the vehicle according to who wins each game. If the pedestrian wins, the experimenter starts closer. If the pedestrian loses, the experimenter starts further away. This feedback process converges to create a high number of *interesting* interactions.

## 3 Results

Data from 27 pedestrians was collected for 20 crossings each. Each crossing took around 1-2 minutes for the pedestrian to complete. [9] No pedestrians were physically hit during the study, but three emergency stops were made using the DMH. On debriefing the subjects in these cases, they all reported deliberately testing the human DMH operator and that they would not have behaved as aggressively if this legally required experimental safety feature was not present.

[9] Video examples of crossings are provided in `https://doi.org/10.5281/zenodo.19818131`

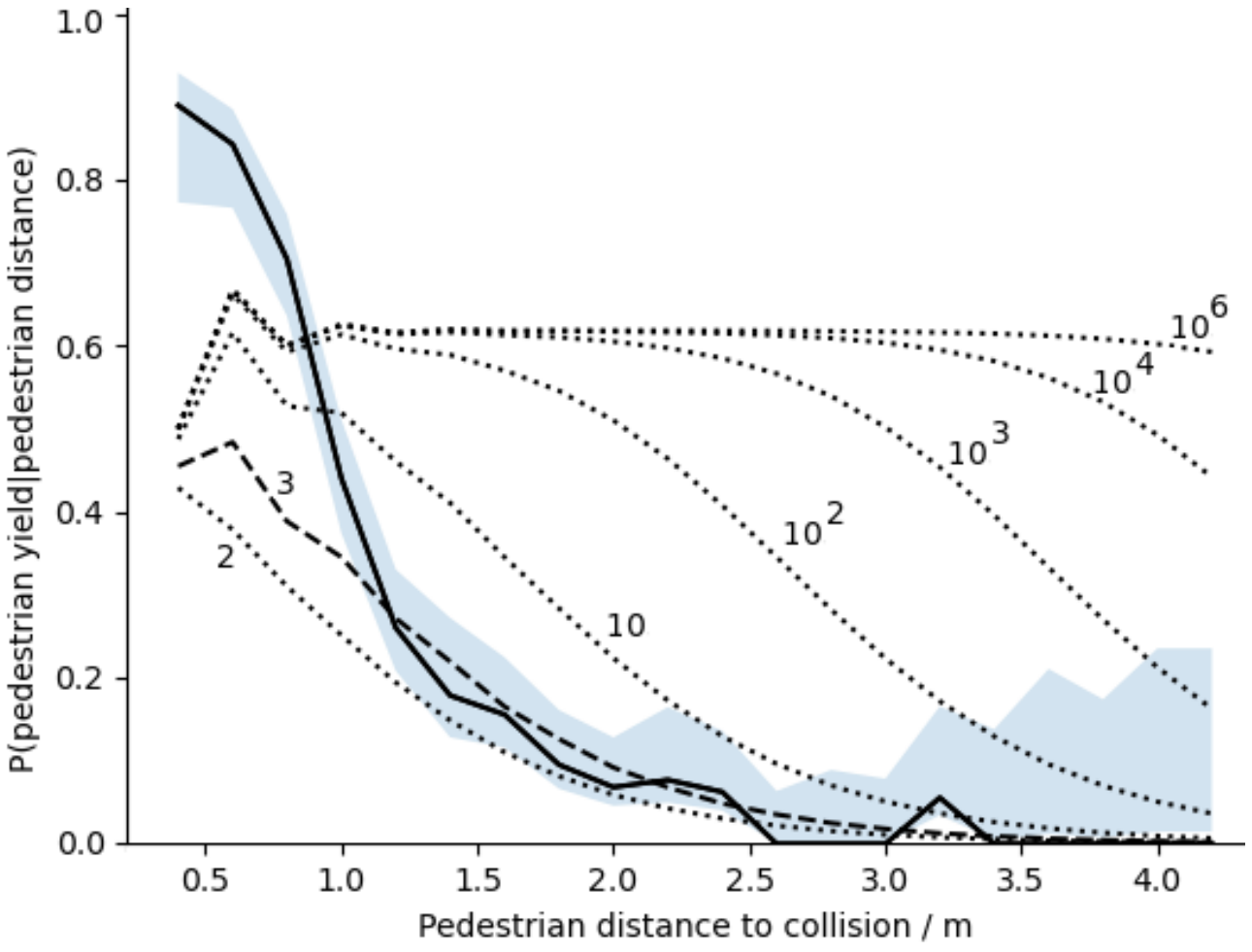

**Fig. 7.** Results. Solid line = empirical mean pedestrian yield probability at distance to collision; gray error band = 1 standard deviation Beta posterior from flat prior. Dashed line = best fit model, $U_{crash} = 3$ seconds. Dotted lines = alternative models with $U_{crash}$ =2, 10, 100, 1,000, 10,000 and 1,000,000 seconds.

It was observed that pedestrians use around the first 2m of their route to accelerate up to their full walking speed, independently of the car position or speed, so the first 2m of each crossing data were removed. When the pedestrian reaches the final observed box, we filter out all subsequent data points of them in that box, so that a maximum of one action is observed there. This is to avoid biasing the data with additional yields after a pedestrian stops to allow the vehicle to pass then has to wait there for it to do so. The data is filtered to include only *interesting* states i.e. where a conflict is present so a game must be played. There are 572 interesting data points, from 5773 total recorded data points, showing that the experimenter feedback protocol and quantization are working well to produce data points of interest in 10% of recorded data. 57% (± 8% per subject) of these interesting states were ultimately won by the pedestrian and 43% by the vehicle. 150 (26%) of the interesting states result in the pedestrian immediately choosing *SLOW*.

The interesting state data is then fit to the model, assuming that both players perceive both of their their times to collision as being equal under the model's quantization, and that they play optimally. Fig. 7 shows the probability of the pedestrian yielding given that a game is played with the pedestrian at a particular distance. We fit the Sequential Chicken model to this data by varying $U_{crash}$, to obtain the best fit at $U_{crash} = 3$ seconds shown in the dotted line. Fig. 8 displays the same results as Fig. 7 but as cumulative probabilities of the pedestrian reaching each distance without having yet performed a yield, assuming that the car also had not yet yielded. This is less sensitive to the box quantization.

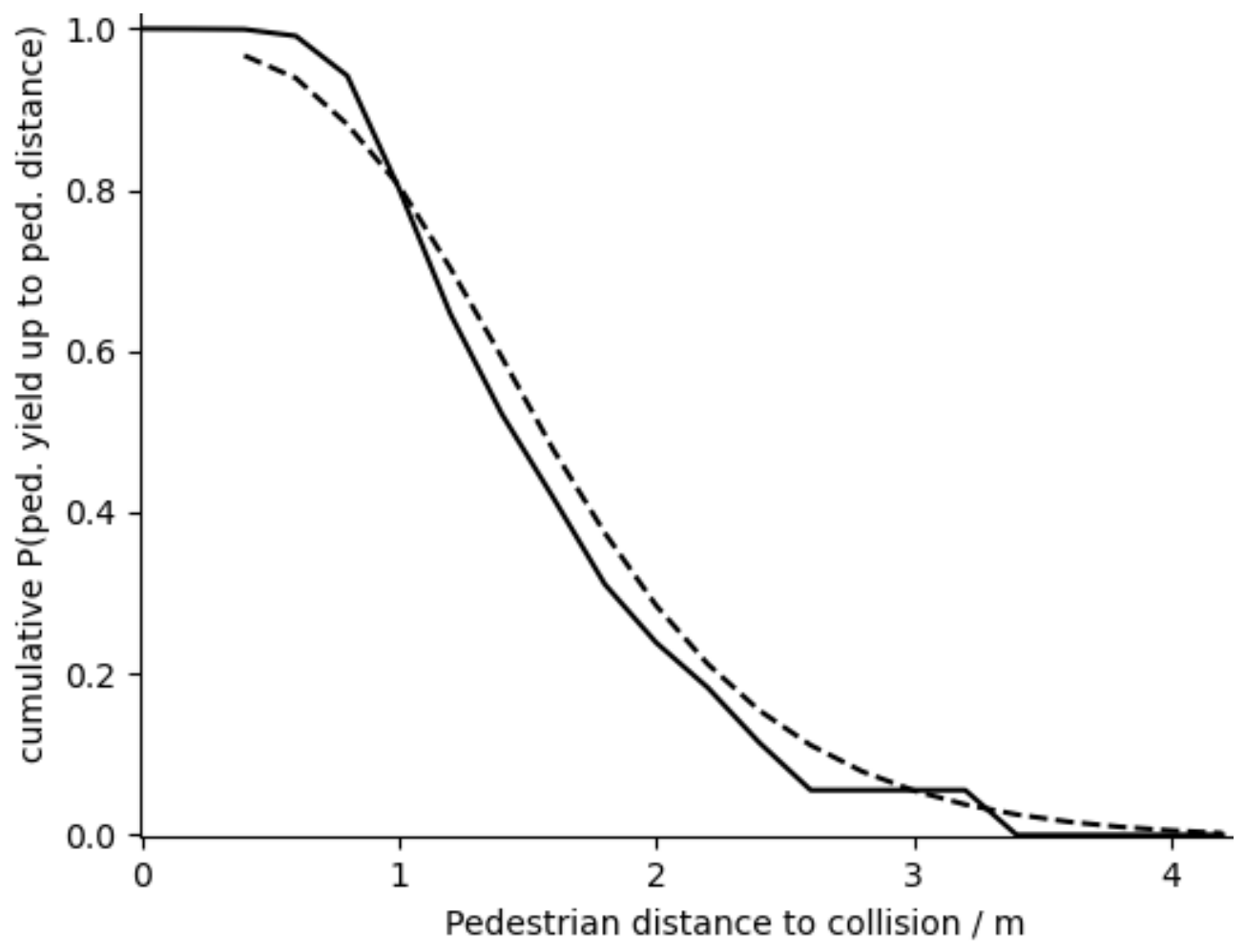

**Fig. 8.** Empirical results (solid) and best fit model (dashed, $U_{crash} = 3$) for cumulative probability pedestrians reaching distances without yielding, if no car yield occurs either.

## 4 Discussion

Fig. 8 shows that pedestrians exhibit sigmoidal increase in their cumulative probability of yielding as the distance to collision decreases rising from around zero at 4m to almost unity, and this pattern is closely predicted by the Sequential Chicken model. This close fit is however obtained under artificial safety requirements which slow down both agents and prevent any actual collisions, and model requirements which restrict them to linear motion. This provides evidence that pedestrians behave according to the model at least in this type of environment.

The implied $U_{crash} = 3$ seconds seems surprisingly small, as in real interactions with traffic we would expect humans to assign much larger values to an actual collision, such as millions of seconds. Safety limitations restricted the speed of both the vehicle and pedestrians to below real-world cases, which (by design) would make physical collision very unlikely to cause actual harm. Subjects were also briefed that the experimenter can intervene to prevent physical harm, and while this defines their task as a failure it may have an unavoidable influence on their behavior. Quantisation may also play a role here: the model makes this fit by assigning $U_{crash}$ to be the penalty of both agents both being in the same box, which can include the possibility of a proxemic 'near miss' penalty as well as an actual collision.

Apart from the three debriefed DMH incidents, almost all pedestrians who reached a game in the collision box then yielded. The model makes no predictions here, as it considers the negative penalty to have already occurred. This might be interpreted as the pedestrian receiving a small proxemic penalty and then intuitively freezing due to surprise discomfort as suggested by proxemic models [11], [23]. Fig. 7 shows that at distances 4m to 1m, Sequential Chicken is a good fit to the pedestrian's *immediate* behavior, but below 1m the model suggests the probability should converge to $\frac{1}{2}$ while the data appears to converge to 1. This discrepancy may be due to actual deviation from game-theoretic optimality in the human behavior as observable in Fig. 9; or to types of quantizations used to fit the discrete model (in particular: to restrict to finite solutions, the model does not include actions to move slower than one box per tick, so quantization needs to make this rare). However Fig. 8 shows that the discrepancy is unimportant in practice because it is very rare for the pedestrian to get into these states – they will usually have yielded much earlier even if the car has not.

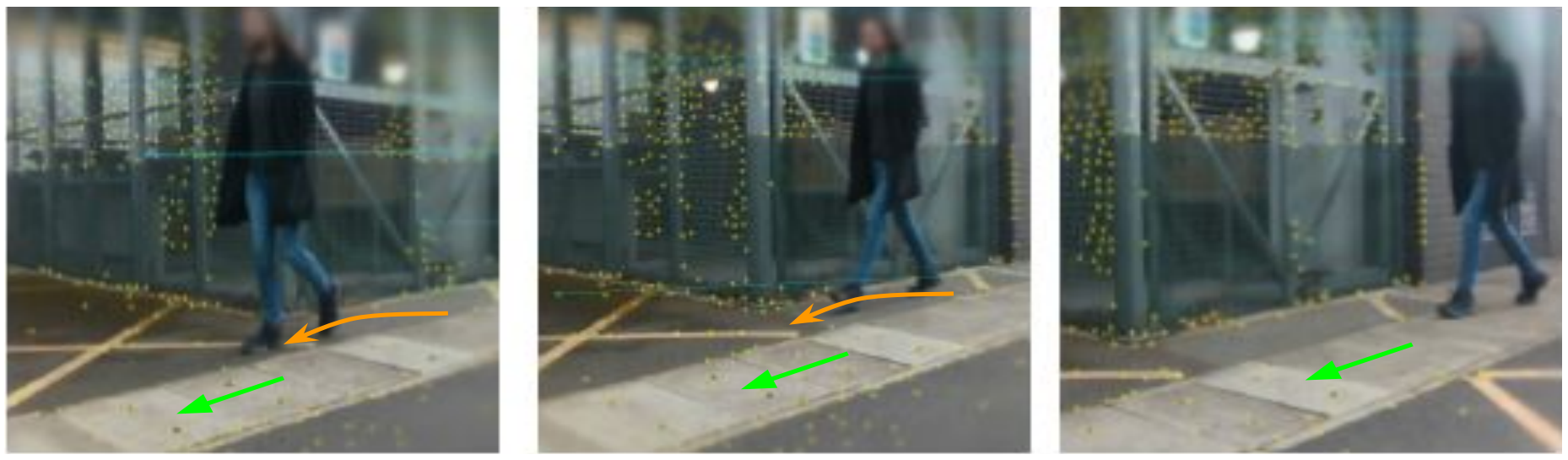

**Fig. 9.** Example of a human participant deviating (orange trajectory) from the prescribed experimental path (green), moving from right to left.

As with all modeling, it may well be possible to find better-fitting models to the data than the proposed Sequential Chicken model. Data from the experiment is available for others to test against new models. The current model constrains both agents to move linearly, as in many road-crossing cases, while in other real life cases such as small delivery vehicles in pedestrianised areas, it is often also possible to swerve. Some models suggest game theory to make choices between curved paths and could be tested here [24], [25], [26], [27], [28]. The particular quantizations used were selected to balance the need to actually have data with both agents are in the same box; with the need for there to be enough boxes to see the shape of the resulting curves. If very small boxes were used, and the agents knew each others positions and speeds exactly, then no game would ever be played because one agent would always be slightly closer than the other so should win deterministically. In reality, both agents have some uncertainty in their perception of the other's position, creating the game theoretic conditions. It is possible that the hand-optimized box quantizations used here may be approximations to some more rigorous probabilistic model, with the box sizes relating to perceptual uncertainties as has been separately modeled [29].

Sequential Chicken has been extended to cases of two agents of different sizes or strengths, which could be fit [4]. A stronger agent like the car, when facing a weaker pedestrian, becomes less likely to yield in the equilibria, while the pedestrian's yield probabilities stay the same. This results in increased probabilities of the stronger agent winning the interaction. The vehicle's own actions in the experiment were programmed using the game theory model, assuming a particular value for $U_{crash}$. In the present experiment, the vehicle's actions were generated using the game-theoretic model with a predefined $U_{crash}$, which necessarily differed from the value later inferred from pedestrian behavior. Future iterations could adapt the vehicle's controller to better align with inferred parameters, though in practice both pedestrians and drivers operate under uncertainty about each other's utilities. Modeling such mutual uncertainty in $U_{crash}$ presents an additional avenue for refinement.

The experimental setting is necessarily artificial, due to the very high safety requirements imposed by human experimental ethics law. VR versions of the experiment were previously conducted [7] [8] which allowed more realistic motions due to the possibility of actual collisions occurring, and the present study provides the first real-world validation of the VR results. Designing experimental protocols to study near-collisions whilst guaranteeing no actual collisions has been found to be very challenging, and the present study serves an example of this general trade-off between realism and safety in experiment design, and of protocols that could be useful for resolving it in similar studies. Building on this first validation, future work could use the current controller to create and analyze data from non-orthogonal vehicle paths, in more natural environments such as delivery robots meeting non-experimental subject pedestrian members of the public. The current analysis assumes a single $U_{crash}$ penalty when the two agents are in the same box, which may conflate aspects of actual physical collision with close but non-contact proxemic space invasion. Future models could separate these effects by introducing graded proxemic penalties (e.g., at intimate and personal distance ranges) alongside distinct, larger penalties for actual collision. Additionally, recent kinematic models suggesting radially asymmetric proxemic zones may offer further refinement of the Sequential Chicken framework [11].

## 5 Conclusion

Pedestrian behaviour in interactions with a physical autonomous vehicle in a controlled experimental setting can be explained by the Sequential Chicken model. Across strategically relevant interaction states, pedestrian yielding behavior exhibited the sigmoidal structure predicted by the model. The relatively small fitted value of $U_{crash}$ may reflect participants' confidence that the onboard safety operator would intervene to prevent a collision. By demonstrating the model in a live physical setting without physical collisions, this study provides proof-of-concept evidence that its strategic formulation can extend beyond simulated environments. Future research may explore richer motion dynamics and more complex and realistic social contexts to further assess the robustness and generalizability of these interaction mechanisms.